\documentclass[hidelinks,onefignum,onetabnum]{siamart220329}

\usepackage[utf8]{inputenc}
\usepackage{bm}
\usepackage{mathrsfs}
\usepackage{amsmath}
\usepackage{amsfonts}
\usepackage{multirow}
\usepackage{dsfont}
\usepackage{caption}
\usepackage{subfigure}
\usepackage{enumitem}

\usepackage[english]{babel}
\makeatletter
\newcommand{\mylabel}[2]{#2\def\@currentlabel{#2}\label{#1}}
\makeatother

\newcommand{\real}{\mathbb{R}}

\newcommand{\R}{\mathbb{R}}

\newcommand{\mcB}{\mathcal{B}}

\newcommand{\mcD}{\mathcal{D}}

\newcommand{\mcG}{\mathcal{G}}

\usepackage{xcolor}

\def\O{\mathcal{O}}

\def\omg{{\Omega}}

\def \bb{\bm{b}}

\def \fb{\bm{f}}

\def \gb{\bm{g}}
\def \ub{\bm{u}}

\def \wb{\bm{w}}
\def \vb{\bm{v}}
\def \xb{\bm{x}}

\def \rb{\bm{r}}

\def \qb{\bm{q}}

\def \yb{\bm{y}}

\def \eb{\bm{e}}

\def \xib{{\boldsymbol\xi}}
\def \etab{{\boldsymbol\eta}}

\def \kb{\bm{k}}

\def \Wb{\bm{W}}

\newcommand{\vertii}[1]{{\left\vert\left\vert #1
    \right\vert\right\vert}}

\usepackage{lineno}

\usepackage{lipsum}
\usepackage{amsfonts}
\usepackage{graphicx}
\usepackage{epstopdf}
\usepackage{algorithmic}
\ifpdf
  \DeclareGraphicsExtensions{.eps,.pdf,.png,.jpg}
\else
  \DeclareGraphicsExtensions{.eps}
\fi

\newsiamremark{remark}{Remark}
\newsiamremark{hypothesis}{Hypothesis}
\crefname{hypothesis}{Hypothesis}{Hypotheses}
\newsiamthm{claim}{Claim}

\headers{Monotone Peridynamic Neural Operator}{J. Wang, X. Tian, Z. Zhang, S. Silling, S. Jafarzadeh and Y. Yu}

\title{Monotone Peridynamic Neural Operator for Nonlinear Material Modeling with Conditionally Unique Solutions\thanks{Submitted to the editors DATE.
\funding{J. Wang and Y. Yu were funded by the National Institute of Health under award 1R01GM157589-01. X.~Tian was supported in part by NSF DMS-2240180 and the Alfred P. Sloan Fellowship. S. Jafarzadeh was funded by the AFOSR grant FA9550-22-1-0197. Portions of this research were conducted on Lehigh University's Research Computing infrastructure, partially supported by NSF Award 2019035.}}}

\author{
Jihong Wang\thanks{Department of Mathematics, Lehigh University, Bethlehem, PA, USA.}
\and Xiaochuan Tian\thanks{Department of Mathematics, University of California, San Diego, CA, USA.}
\and Zhongqiang Zhang\thanks{Department of Mathematical Sciences, Worcester Polytechnic Institute, Worcester, MA, USA.}
\and Stewart Silling\thanks{Center for Computing Research, Sandia National Laboratories, Albuquerque, NM, USA.}
\and Siavash Jafarzadeh\footnotemark[2] 
\and Yue Yu\footnotemark[2] \thanks{Corresponding author. (\email{yuy214@lehigh.edu}).}
}

\usepackage{amsopn}

\begin{document}

\maketitle
\begin{abstract}
Data-driven methods have emerged as powerful tools for modeling the responses of complex nonlinear materials directly from experimental measurements. Among these methods, the data-driven constitutive models present advantages in physical interpretability and generalizability across different boundary conditions/domain settings. However, the well-posedness of these learned models is generally not guaranteed a priori, which makes the models prone to non-physical solutions in downstream simulation tasks.
In this study, we introduce monotone peridynamic neural operator (MPNO), a novel data-driven nonlocal constitutive model learning approach based on neural operators.
Our approach learns a nonlocal kernel together with a nonlinear constitutive relation, while ensuring solution uniqueness through a monotone gradient network. This architectural constraint on gradient induces convexity of the learnt energy density function, thereby guaranteeing solution uniqueness of MPNO in small deformation regimes. 
To validate our approach, we evaluate MPNO's performance on both synthetic and real-world datasets. On synthetic datasets with manufactured kernel and constitutive relation, we show that the learnt model converges to the ground-truth as the measurement grid size decreases both theoretically and numerically. 
Additionally, 
our MPNO exhibits superior generalization capabilities than the conventional neural networks: it yields smaller displacement solution errors in down-stream tasks with new and unseen loadings. Finally, we showcase the practical utility of our approach through applications in learning a homogenized model from molecular dynamics data, highlighting its expressivity and robustness in real-world scenarios.
\end{abstract}

\begin{keywords}
Neural operators, Data-driven physics modeling, Well-posedness,  Peridynamics, Nonlocal models
\end{keywords}

\begin{MSCcodes}
68Q25, 68R10, 68U05
\end{MSCcodes}



\section{Introduction}


Learning complex physical systems from data has become increasingly prevalent in scientific and engineering fields \cite{karniadakis2021physics,liu2024deep,ghaboussi1991knowledge,carleo2019machine,zhang2018deep,liu2023ino,cai2022physics,pfau2020ab,ruthotto2020deep}, from biomedical engineering \cite{ma2017pre,linka2023automated,zheng2022data,tac2022data}, turbulence modeling \cite{xu2022pde,wang2017physics,oommen2024integrating,bai2017data} weather forecasting \cite{kashinath2021physics,liu2023clawno}, to material modeling \cite{trask2024unsupervised,geng2025end,bhattacharya2025learning,goswami2021physics}. In many cases, the underlying governing equations remain unknown, and must be uncovered from data. Taking material modeling for instance, the governing equation should be described as a constitutive model, mapping from displacement (or strain) fields to the corresponding body force (or stress) fields. For many decades, constitutive models with pre-defined PDEs have been commonly employed to characterize material response. However, the descriptive power of such predefined constitutive models often relies on the prior physics knowledge, which restricts their applicability and robustness in many real-world applications, such as the new material characterization from experimental measurements.

To address the limitation of pre-defined constitutive models, data-driven constitutive models have been developed, e.g., gradient sensitivity methods \cite{akerson2024learning}, physics-informed machine learning \cite{haghighat2023constitutive}, probabilistic machine learning \cite{fuhg2022physics}, deep learning \cite{linka2023automated,tac2022data,liu2019deep}, and operator learning \cite{jafarzadeh2024peridynamic,bhattacharya2025learning}. 
In contrast with the equation-free learning approach \cite{kevrekidis2004equation} which often parameterizes the mapping from force loadings/boundary conditions to the corresponding displacement field directly \cite{kevrekidis2004equation} as a (hidden) PDE solution operator, the data-driven constitutive modeling approach aims to discover unknown constitutive equations. As a result, while the equation-free approach is often restricted to fixed loading/domain settings, the data-driven constitutive equation is generalize to new and unseen loading/boundary conditions as well as spatial domains in down-stream modeling tasks \cite{jafarzadeh2024peridynamic}. Additionally, the data-driven constitutive model also provides interpretable physical insights of the material responses, e.g., in characterizing the material heterogeneity \cite{jafarzadeh2024heterogeneous} and nonlinearity \cite{he2021manifold}. These properties make data-driven constitutive modeling approaches an actively developing field \cite{fuhg2024review}. Among these approaches, we mention the peridynamic neural operator (PNO) \cite{jafarzadeh2024peridynamic}, which provides the constitutive equation as a peridynamic model \cite{silling_2010} in a special form of nonlocal neural operators \cite{li2020fourier,li2020neural}, with Galilean invariance, frame invariance, and momentum balance laws guaranteed. On real-world modeling tasks including homogenized model learning \cite{you2021md} and material heterogeneity characterization \cite{jafarzadeh2024heterogeneous}, PNO was employed to learn directly from displacement field–loading field data pairs, and has shown improved accuracy against pre-defined PDE models and capability in discovering nonlinear, anisotropic, and heterogeneous responses.

However, applying PNOs and other data-driven constitutive models in down-stream modeling tasks may not lead to well-posed models. 
In \cite{tac2022data}, Tac et al. proposed a neural ordinary differential equation-based approach, to obtain the strain energy potential in terms of the sums of convex non-decreasing functions of invariants by fitting stress-strain curves. As such, poly-convexity of the strain energy is automatically satisfied, guaranteeing the existence of physically realistic solutions. However, this approach requires prior knowledge on the involved invariant terms, and it can only learn from stress-strain curves. That means, it can be difficult to handle material heterogeneity and more general types of training data beyond stress-strain curves, such as the full-field spatial measurements from experiments \cite{fitzpatrick2022ex} or high-fidelity simulations \cite{you2021md}.

In this study, we aim to address the question of how to obtain from full-field displacement measurements a constitutive model that captures complex material responses with guaranteed well-posedness. To handle the full-field displacement measurements, we propose to leverage the PNO architecture \cite{jafarzadeh2024peridynamic,jafarzadeh2024heterogeneous}, which captures the spatial measurements by learning a surrogate mapping between function spaces while preserving basic physics requirements. To enforce the well-posedness of learnt model, two developments are desired: a uniqueness condition for nonlinear peridynamics, and a corresponding neural network architecture satisfying this condition. To this end, we first conduct a rigorous analysis of the condition under which nonlinear peridynamics admits a unique solution, and then show that learning a monotone gradient network between the bond stretch and the micropotential function would guarantee this condition in a small deformation strain. 
We summarize below our main contributions:
\begin{itemize}
\item We propose the Monotone Peridynamic Neural Operator (MPNO), a novel nonlocal neural operator-based constitutive law formulation with guaranteed model well-posedness. The key idea is to parameterize the constitutive relation as a monotone gradient network, which automatically satisfies the convexity of the strain energy for nonlinear peridynamics.
\item We have established, for the first time  a sufficient condition for the solution uniqueness for the nonlinear bond-based peridynamics corresponding to small relative displacements. This result not only serves as the theoretical foundation of our MPNO, but also leads to a new rigorous study of nonlocal mechanics model with well-posedness similar to their PDEs counterparts.
\item As a result, a nonlocal homogeneous constitutive model is obtained in the form of bond-based peridynamics. In contrast with classical PDE-based models, our MPNO can capture nonlinear material responses without prior expert-constructed knowledge on the invariant terms, while guaranteeing solution uniqueness in the small deformation regime.
\item A learning algorithm is proposed to infer the constitutive relation simultaneously with a kernel function from data. We verify the efficacy of MPNO and this learning algorithm on synthetic datasets, and for the first time demonstrating that the learnt model converges to the ground-truth as the measurement grid size decreases. Moreover, in down-stream tasks with new and unseen loading scenarios, the solution from our model also converges with the grid size, highlighting the advantage of learning a well-posed model. 
\end{itemize}

The remainder of this paper is organized as follows. Section \ref{sec:background} introduces the peridynamic theory and the particular class of material models considered in this study, followed by an overview of the peridynamic neural operator (PNO). In Section \ref{sec:method}, we present the detailed methodology of MPNO. Specifically, we analyze the uniqueness of the continuum model and incorporate the monotone gradient network (MGN) into the PNO framework to ensure this property in small deformation regimes. A learning algorithm is summarized in Algorithm \ref{alg:1}. To evaluate the proposed model and algorithm, Section \ref{sec:experiments} considers a synthetic dataset, focusing on the convergence behavior and the properties of MPNO with guaranteeing uniqueness. In Section \ref{sec:app}, we apply MPNO to learn the continuum constitutive model for a molecular dynamics dataset, highlighting its practical applicability. Finally, Section \ref{sec:conclusions} provides a summary.

\section{Background}\label{sec:background}

In this section, we provide a brief introduction to peridynamic theory, particularly on the bond-based peridynamic model used in this work. Then, we present an overview of the peridynamic neural operator framework \cite{jafarzadeh2024peridynamic}. 

\subsection{Peridynamic model}\label{subsec-peri}
Peridynamics reformulates continuum mechanics by employing integral equations instead of classical differential equations \cite{silling2000reformulation,seleson2009peridynamics,parks2008implementing,zimmermann2005continuum,emmrich2007analysis,du2011mathematical,bobaru2016handbook}. This nonlocal framework introduces a characteristic length scale $\delta$, known as the horizon, which defines the range of interactions between material points.
Peridynamic equations of motion can be categorized into two main types: {\emph{bond-based}} and {\emph{state-based}}. In the bond-based formulation, the interaction between any pair of material points depends only on their relative displacement, independent of other points within their interaction domain. In contrast, the state-based formulation accounts for the collective influence of all points within the interaction domain. This study uses bond-based peridynamic material modeling, because of its greater simplicity.
Given a domain $\Omega\subset\R^d$, called the {\emph{reference configuration}}, the motion of a material point $\xb\in\Omega$ at time $t$ is governed by the following bond-based peridynamic equation:
\begin{equation}\label{eq:bbpd}
    \rho(\xb)\ddot\ub(\xb,t)= \int_{{B_\delta(\xb)}}\fb(\ub(\qb, t) - \ub(\xb, t), \qb-\xb)\;d\qb+\bb(\xb,t)\text{ ,}\quad (\xb,t)\in\omg\times[0,T]\text{ ,}
\end{equation}
where $\ub$ is the displacement vector field, $\ddot\ub$ is the acceleration, $\bb$ is a prescribed body force density field, $\rho$ is the mass density in the reference configuration, and $\fb$ is the  {\emph{pairwise force density}}, whose value is the force vector (per unit volume squared) that the particle $\qb$ exerts on the particle $\xb$. The interaction between $\xb$ and $\qb$ is called a {\emph{bond}}. The relative position of the endpoints of the bond vector in the reference configuration is denoted by $\xib$ and their relative displacement by $\etab$:
\begin{equation}
\xib = \qb-\xb,~ \etab = \ub(\qb, t)-\ub(\xb,t).
\end{equation}
The {\emph{bond stretch}} is defined as $\lambda = |\xib + \etab|/ |\xib|$, which is the ratio of the deformed bond length to its undeformed length. Here, we note that two material points should not cross each other, and hence $\lambda>0$. 
It is assumed that the material is {\emph{microelastic}}, which means that it cannot sustain permanent deformation. The stored energy in a bond under deformation is called the {\emph{micropotential}} $w(\lambda, \boldsymbol{\xi})$, which has dimensions of energy per unit volume squared and is non-negative.
The energy per unit volume in the body at a given point (i.e., the local strain energy density) is found from
\begin{equation}\label{eq:energy_density}
    W(\xb) = \frac12\int_{{B_\delta(\bm{0})}}w(\lambda, \xib)\,d\xib
\end{equation}
where $B_\delta(\bm{0})$ is the neighborhood of radius $\delta$ centered at $\bm{0}$, known as the {\emph{family}} of $\xb$.
The factor of $1/2$ appears because each endpoint of a bond “owns” only half the energy in the bond. If a body is composed of a microelastic material, work done on it by external forces is stored in recoverable form in much the same way as in the classical theory of elasticity.

In a microelastic material the pairwise force function is derived from $w$:
\begin{equation}
    \fb(\etab,\xib) = \frac{\partial w}{\partial\etab}=\frac{\partial w}{\partial \lambda}(\lambda, \xib)\frac{\partial \lambda}{\partial \etab} = \frac{\partial w}{\partial \lambda}(\lambda,\xib)\frac{1}{|\xib|} \frac{\xib+\etab}{|\xib+\etab|}, \quad \forall \xib, \etab.
\end{equation}
This shows that that the force vector in a bond is parallel to the current relative position vector between the two particles. It also shows that the force vector exerted by any particle $\qb$ on $\xb$ equals minus the force vector that $\xb$ exerts on $\qb$. Conservation of angular and linear momenta is therefore automatically satisfied by a microelastic material model. Also, since $\partial w/\partial \lambda$ is invariant with respect to rigid rotation of a bond, the requirement of objectivity is trivially satisfied.

It is assumed here that $w(\lambda, \xib)$ can be written in the separable form and  we denote
\begin{equation}\label{eq:w}
 \frac{\partial w}{\partial \lambda}(\lambda,\xib)\frac{1}{|\xib|} :=g(\lambda)k(\xib), \, \text{where }k(\xib)\geq 0.
\end{equation}
This asserts that all the bond forces depend on the bond deformation in essentially the same way, but the forces could be lesser or greater for different bond lengths.
This assumption is not completely general, but it provides a substantial simplification and allows the most important features of material response to be modeled accurately.
The corresponding form of the equation of motion is then given by
\begin{equation}\label{eq:model}
     \rho(\xb)\ddot\ub(\xb,t)=\int_{B_{\delta}(\bm{0})} g(\lambda)k(\xib) \frac{\xib+\etab}{|\xib+\etab|} d\xib + \bb(\xb,t), \quad (\xb,t)\in\omg\times[0,T]\text{ ,}
\end{equation}
with the boundary data is prescribed in the nonlocal boundary layer $\Omega_I=\{\xb|\xb\in\real^d \backslash \omg,\,\text{dist}(\xb,\omg) < \delta\}$, i.e.,
\begin{equation} \label{eq:bc}
\ub(\xb,t)=\ub_{BC}(\xb,t),\quad (\xb,t)\in\Omega_I\times[0,T].
\end{equation}



\subsection{Peridynamic neural operators}\label{subsec:pno}

Peridynamic neural operator \cite{jafarzadeh2024peridynamic,jafarzadeh2024heterogeneous} 
employs neural operators to parameterize the constitutive equation as a function-to-function mapping. 
Denote the input and output function spaces as $\mathcal{U}=\mathcal{U}(\Omega;\R^{d_u})$ and $\mcB=\mathcal{B}(\Omega;\R^{d_b})$, respectively.
Neural operators construct a surrogate mapping $\mathcal{G}:\mathcal{U}\rightarrow\mathcal{B}$ that transforms an input function $\ub(\xb)\in\mathcal{U}$ to an output function $\bb(\xb)\in\mathcal{B}$ through resolution-independent nonlocal integral transformations. 
Then, the PNO is proposed based on the following state-based peridynamic model
\begin{equation}\label{eqn-Gapprox}
  \rho(\xb)\ddot\ub(\xb,t) = \mcG[\ub](\xb,t) +\bb(\xb,t),
\end{equation}
where the operator $\mcG$ is a nonlocal operator in the form of state-based peridynamics \cite{bobaru2016handbook}.
In PNOs, $\mcG$ is approximated with a neural operator $\mcG^{NN}[\ub;\theta]$, and the neural network parameter $\theta$ is inferred from dataset of displacement/loading function pairs $\{(\ub^{(i)}(\xb,t),\bb^{(i)}(\xb,t))\}_{i=1}^N$ by minimizing the discrepancy between the operator output $\mcG^{NN}[\ub^{(i)};\theta]$ and the true physical residual $\rho\ddot{\ub}^{(i)} - \bb^{(i)}$ over a set of discretized grid points $\chi = \{\xb_1,...,\xb_M\} \subset \Omega$ and temporal steps $t_n = n\Delta t$. Once trained, the model can predict mechanical responses for new loading conditions, maintaining physical consistency while demonstrating improved generalization across varying geometries and boundary conditions. In the following, we focus on the learning problem, and consider the (quasi)static model, i.e., $\ub(\xb,t)=\ub(\xb)$. 

\section{Monotone peridynamic neural operators (MPNOs)} \label{sec:method} 
Even though nonlinear bond-based peridynamics are increasingly popular, its well-posedness remains largely unexplored.  
Herein, we establish the uniqueness analysis of \eqref{eq:model}-\eqref{eq:bc} and propose a corresponding learning framework that guarantees uniqueness. Specifically, we first conduct a rigorous analysis of a sufficient condition under which the bond-based peridynamic model admits a unique solution under small deformation. Based on this theoretical foundation, we encode this condition directly into the neural operator architecture, i.e., in the operator $\mcG$. Finally, we present a detailed learning algorithm designed to train models that inherently guarantee uniqueness, bridging the gap between theory and practice in data-driven constitutive modeling.

\subsection{Uniqueness: construction of convex energy functions}\label{subsec:uni}

In this subsection, we establish uniqueness of the solution under suitable assumptions. 
We then consider the following energy functional for the model \eqref{eq:model}-\eqref{eq:bc}:
\begin{equation}\label{eq:energy}
E[\ub] =\frac{1}{2}  \int_{\omg\cup\omg_I}  \int_{B_\delta(\bm{0})\cap (\omg\cup\omg_I - \xb)}w(\lambda, \xib) d\xib d\xb + \int_\omg  \bb(\xb) \ub d\xb
\end{equation}
with the Dirichlet condition. Here, $\omg\cup\omg_I - \xb: = \{ \xib\in\R^d: \xb + \xib \in \omg\cup\omg_I \}$.
A direct variational calculation yields the corresponding Euler–Lagrange equation:
\begin{equation}\label{eqn:fullmodel}
\int_{B_{\delta}(\bm{0})} g(\lambda)k(\xib) \frac{\xib+\etab}{|\xib+\etab|} d\xib + \bb(\xb) = {\bf 0}\quad \xb\in\omg.
\end{equation}

When we assume small deformations, the following linearization of $\lambda$ holds: 
\begin{eqnarray}\label{eq:lambda_linear}
\lambda = |\xib + \etab|/ |\xib| \approx \frac{|\xib| + \xib\cdot \etab/|\xib|}{|\xib|} = 1 + \frac{\xib\cdot \etab}{|\xib|^2}.
\end{eqnarray}

\begin{definition}
Two deformations $\wb$ and $\tilde\wb$ are {\underline{equivalent}} if $\lambda=\tilde\lambda$ for all bonds.
\end{definition}

If $\wb$ and $\tilde\wb$ are equivalent, they differ from each other by a rigid translation, rigid rotation, or some combination of the two.



\begin{theorem}\label{lemma}
Under the small deformation assumption given in \eqref{eq:lambda_linear}, if the function $w(\lambda, \xib)$ is strictly convex with respect to $\lambda \in (0,\infty)$, then any two minimizers of the energy functional $E[\ub]$ defined in \eqref{eq:energy} are equivalent. 
\end{theorem}
\begin{proof}
Under the small deformation assumption, the energy is written as
\[
E[\ub] =\frac{1}{2} \int_{\omg\cup\omg_I}  \int_{B_\delta(\bm{0})\cap (\omg\cup\omg_I - \xb)}  w\left( 1 + \frac{\xib\cdot \etab}{|\xib|^2}, \xib\right)  d\xib d\xb + \int_\omg  \bb(\xb) d\xb. 
\]
It can be readily verified that if $w(\cdot, \xib )$ is strictly convex in its first argument for every $\xib$, then $E$ is strictly convex, i.e., 
\[
E(c \ub  + (1-c) \vb ) \leq c E(\ub) + (1-c) E(\vb), \quad \forall c\in [0,1],
\]
with equality holding if and only if $\ub$ and $\vb$ are equivalent. We now show that for strictly convex $w$, any minimizer of $E$ must be unique. Suppose $\wb$ and $\tilde{\wb}$ are both minimizers, i.e.,
\[
m = \inf E(\ub) = E(\wb) = E(\tilde{\wb}). 
 \] 
Then by convexity, 
\[
E\left(\frac{1}{2} ( \wb +\tilde{\wb})\right) \leq \frac{1}{2} (E(\wb) + E(\tilde{\wb})) = m. 
 \]
Since $m$ is the minimal value, this inequality must be an equality. By strict convexity of  $E$, it follows that $\lambda=\tilde\lambda$ and therefore $\wb$ and $\tilde{\wb}$ are equivalent.
\end{proof}


\begin{remark}
\label{rmk:monotonicity}
Let $A\subset \R^n$ be an open and convex set and $f: A \to \R$ be a convex and differentiable function. 
It is a classical result (see e.g., \cite{rockafellar1997convex}) that the gradient map $\nabla f :  A \to \R^n$ is monotone in the sense that 
$
(\nabla f (\xb) - \nabla f(\yb)) \cdot (\xb - \yb) \geq 0, \quad \forall \xb, \yb \in A. 
$
\end{remark}

\subsection{Proposed architecture}\label{subsec:mpno}
Following the idea of PNO in \cite{jafarzadeh2024peridynamic}, we now construct a surrogate operator $\mcG:\mathcal{U}\rightarrow\mathcal{B}$ that maps the input function $\ub(\xb)$ to the output function $\bb(\xb)$ in \eqref{eq:model}, with the uniqueness condition in Theorem \ref{lemma}. 

First, we parameterize the nonlinear function $g(\lambda)$ and kernel function $k(\xib)$ by two neural networks $g^{NN}$ and $k^{NN}$, respectively, i.e., we have
\begin{equation} \label{eq:G}
\mcG[\ub](\xb) \approx \mcG^{NN}[\ub;\theta](\xb):=\int_{{B_\delta(\bm{0})}} g_{\theta_g}^{NN}(\lambda) k_{\theta_k}^{NN}(\xib) \frac{\xib+\etab}{|\xib+\etab|} \;d\xib.
\end{equation}
Here, $\theta_g$ and $\theta_k$ represent the trainable parameters of $g^{NN}$ and $k^{NN}$, respectively, and $\theta=\{\theta_g,\theta_k\}$. To guarantee the uniqueness of the solution, we incorporate the the convex property of the function $w(\lambda, \xib)$ using \Cref{rmk:monotonicity}. 
That means, if $g(\lambda)$ is a monotonically increasing function of  $\lambda$, $w(\lambda, \xib)$ will be convex with respect to $\lambda$, and the resultant model has solution uniqueness. 
To this end, we propose to employ the cascaded monotone gradient network (mGradNet-C) \cite{chaudhari2024gradient}, a neural network architecture designed to directly learn the gradients of convex functions:
\begin{eqnarray}
\begin{aligned}
z_0 &= \beta_0 \odot \Wb\lambda + b_0, \\
z_{l} &= \beta_{l} \odot \Wb\lambda + \alpha_{l} \odot \sigma_{l} (z_{l-1}) + b_{l},\, l=1,\cdots,L-1,\\
\text{mGradNet-C}(\lambda) &= \Wb^{\top} \left[ \alpha_L \odot \sigma_L (z_{L-1}) \right] + b_L.
\end{aligned}
\end{eqnarray}
Here, $\lambda$ is the input, $z_l$ is the output of each layer, $\alpha_l, \beta_l$ are nonnegative scaling weights, and $\Wb$ is a weight matrix shared by all $L$ layers. That means, we have $\theta_g=\{\beta_l,\alpha_l,b_l,\Wb\}$. $\sigma_l$ are differentiable, monotonically-increasing elementwise activation functions, which may vary across different layers. The symbol $\odot$ denotes the Hadamard (entrywise) vector product. In our learning framework, we parameterize nonlinear constitutive relation function $g(\lambda)$ using mGradNet-C, and the kernel function $k(\xib)$ using a multi-layer perceptron (MLP) \cite{riedmiller2014multi}, with a ReLU function in the output layer to guarantee that the kernel is non-negative. 




\subsection{Learning algorithm}
Although the prescribed architecture can be readily extended to higher-dimensional domains and dynamic cases, in this work, we focus on 1D and 2D (quasi)static tasks. 
Given discrete observations of function pairs:
\begin{eqnarray*}
\mathcal{D} = \{ (\ub^{(i)},\bb^{(i)})\}_{i=1}^{N} = \{ (\ub^{(i)}(\xb_j), \bb^{(i)}(\xb_j)): j=1,\cdots, N_x\}_{i=1}^{N}, 
\end{eqnarray*}
where $N$ is the number of samples, $\{ (\ub^{(i)},\bb^{(i)})\}$ represent pairs of real-valued continuous functions defined on a bounded, open, and connected domain  $\Omega\subset \R^d$, and $\chi = \{ \xb_j\in\Omega\}$ denotes the set of spatial mesh points.
For each function pair and each bond of $(\xb_k,\xb_j)$ with $\xb_k\in B_{\delta}(\xb_j)\cap \chi$, 
we then compute the corresponding bond stretch $\lambda^{(i)}_{j,k-j}:=|\xb_k-\xb_j+\ub^{(i)}(\xb_k)-\ub^{(i)}(\xb_j)|/|\xb_k-\xb_j|$, original bond $\xib_{j,k-j}:=\xb_k-\xb_j$, and deformed bond $\etab^{(i)}_{j,k-j}:=\ub^{(i)}(\xb_k)-\ub^{(i)}(\xb_j)$. The neural networks $g^{NN}$ and $k^{NN}$ take $\lambda^{(i)}_{j,k-j}$ and $\xib_{j,k-j}$ as their respective inputs. The integral in  \eqref{eq:G} is then approximated using a Riemann sum:
\begin{align}\label{eqn:Riemann_sum}
\mcG_{\Delta x}^{NN}[\ub^{(i)};\theta](\xb_j):=&\underset{\xb_k\in B_{\delta}(\xb_j)\cap \chi}{\sum} g_{\theta_g}^{NN}(\lambda^{(i)}_{j,k-j})k_{\theta_k}^{NN}(\xib_{j,k-j}) \frac{\xib_{j,k-j}+\etab^{(i)}_{j,k-j}}{|\xib_{j,k-j}+\etab^{(i)}_{j,k-j}|}w_{jk}.
\end{align}
Here, $w_{jk}=\Delta x^d$ and $\Delta x$ is the  uniform grid size, and $d$ is the dimension of $\Omega$. 

To obtain the optimal model approximators $g^{NN}(\lambda)$ and $k^{NN}(\xib)$, we split the dataset $\mcD$ into three non-overlapping subsets: 
training, validation, and test, respectively. Then we minimize the loss function with respect to the trainable parameter $\theta=\{{\theta_g},{\theta_k}\}$ on the trianing dataset:
\begin{eqnarray}\label{eq:loss}
 \text{loss}(\theta) := &\frac{1}{N_{train}}\sum_{i= 1}^{N_{train}}\dfrac{\vertii{\mcG_{\Delta x}^{NN}[\ub_{train}^{(i)};\theta]+\bb_{train}^{(i)}}_{l^2}}{\vertii{\bb_{train}^{(i)}}_{l^2}},
\end{eqnarray}
where $\vertii{\cdot}_{l^2}$ denotes the $l^2$ norm over the involved points. Notably, this learning algorithm exhibits resolution independence, as the prediction accuracy remains invariant with respect to the resolution of the input functions.


\subsection{Evaluation metrics} After training, the performance of the model will be evaluated on samples with new and unseen loadings $\mcD_{test}=\{(\ub^{(i)}_{test}(\xb),\bb^{(i)}_{test}(\xb))\}_{i=1}^{N_{test}}$, based on three criteria:

1) Error of $\bb$ (residual of the model equation):
\begin{equation}\label{eqn:Err_b}
E_b:=\frac{1}{N_{test}}\sum_{i=1}^{N_{test}}\dfrac{\vertii{\mcG_{\Delta x}^{NN}[\ub_{test}^{(i)};\theta]+\bb_{test}^{(i)}}_{l^2}}{\vertii{\bb_{test}^{(i)}}_{l^2}}.
\end{equation}
This error evaluates the (relative) residual at test samples.

2) Error of model ($k$ and $g$): 
\begin{align}
&E_k:=\dfrac{\vertii{k^{NN}_{\theta_k}-k}_{l^2(\rho_\xi)}}{\vertii{k}_{l^2(\rho_\xi)}},\,E_g:=\dfrac{\vertii{g^{NN}_{\theta_g}-g}_{l^2(\rho_\lambda)}}{\vertii{g}_{l^2(\rho_\lambda)}},\, E_{gk}:=\dfrac{\vertii{g^{NN}_{\theta_g}k^{NN}_{\theta_k}-gk}_{l^2(\rho_\lambda\rho_\xi)}}{\vertii{gk}_{l^2(\rho_\lambda\rho_\xi)}}.\label{eqn:Err_kg} \hskip -2pt 
\end{align}
Here, $l^2(\rho_\xi)$ and $l^2(\rho_\lambda)$ are the empirical measures in the kernel spaces of $k$ and $g$, respectively. In the following we provide the formal definition of $l^2(\rho_\xi)$ and $l^2(\rho_{\lambda})$. This error evaluates the discrepancy of the learnt model. Intuitively, $g$ and $k$ are learnt with an unsupervised way, and hence obtaining an accurate $g$ and $k$ is a generally more challenging task comparing to obtaining a good regression error $E_b$.

3) Error of $\ub$: 
\begin{equation}\label{eqn:Err_u}
E_u:=\frac{1}{N_{test}}\sum_{i=1}^{N_{test}}\dfrac{\vertii{(\mcG_{\Delta x}^{NN})^{-1}[-\bb_{test}^{(i)};\theta]-\ub_{test}^{(i)}}_{l^2}}{\vertii{\ub_{test}^{(i)}}_{l^2}}.
\end{equation}
Here, $(\mcG_{\Delta x}^{NN})^{-1}[-\bb_{test}^{(i)};\theta]$ denotes the numerical solution of learnt model under a new and unseen loading $\bb_{test}^{(i)}$, and satisfying the Dirichlet boundary condition $\ub(\xb)=\ub^{(i)}_{test}(\xb)$ on the nonlocal boundary $\omg_I$. This error characterizes the performance of learnt model in down-stream material simulation tasks, which may be with a different domain and loading setting outside the training regime. Intuitively, if the model is not well-posed, $(\mcG_{\Delta x}^{NN})^{-1}[-\bb_{test}^{(i)};\theta]$ may have non-physical or even divergent solutions.

\textbf{Empirical measures for $g$ and $k$.} We now introduce two probability measures, $l^2(\rho_{\lambda})$ and $l^2(\rho_{\xi})$, which can be seen as an extension of the kernel exploration measure defined in \cite{lu2022nonparametric,yu2024nonlocal}, to quantify the extent to which the data explores the variables of $g(\lambda)$ and $k(\xi)$. For $\rho_\xi$ we denote 
\begin{equation}\label{eq:L2rho_xi}
   \rho_\xi(d\rb) = \sum_{i=1}^N \sum_{j\in \chi}\sum_{\xb_k\in B_\delta(\xb_j)} \mathds{1}_{\xb_k-\xb_j=\rb}(\rb) \frac{w^{(i)}_{j,k-j}}{|\mathcal{R}|},
\end{equation}
where $\mathds{1}_{A}(\rb)$ is the indicator function of set $A$, that takes the value 1 when $\rb\in A$ and zero otherwise. The weight function
$\displaystyle w^{(i)}_{j,k-j} = |g(\lambda^{(i)}_{j,k})(\xib_{j,k-j}+\etab^{(i)}_{j,k-j})|/|\xib_{j,k-j}+\etab^{(i)}_{j,k-j}|$ with $|\mathcal{R}|$ representing the number of all data points for which $|\xib_{j,k-j}|\leq \delta$.
To define the measure $\rho_{\lambda}$ in a similar way, we first divide the range of the bond stretch $\lambda$ into $N_{\lambda}$ intervals: $I_n = [\lambda_n, \lambda_{n+1}]$, $n=0,1,\cdots, N_{\lambda}-1$. 
The $\rho_\lambda$ measure is then given by
\begin{equation}\label{eq:L2rho_lambda}
   \rho_\lambda(d\lambda) = \sum_{i=1}^N \sum_{j\in \chi}\sum_{\xb_k\in B_\delta(\xb_j)} \mathds{1}_{I_n}(\lambda) \frac{v^{(i)}_{j,k-j}}{|\mathcal{R}|},
\end{equation}
where the weight function is defined by
$\displaystyle v^{(i)}_{j,k-j} = |k(\xib_{j,k-j})(\xib^{(i)}_{j,k-j}+\etab^{(i)}_{j,k-j})|/|\xib^{(i)}_{j,k-j}+\etab^{(i)}_{j,k-j}|$.
In the following, all the model errors reported in the following experiments are measured in the $l^2(\rho)$ norm. Specifically, for the function $k(\xib)$, we use the $l^2(\rho_\xi)$ error, while for the function $g(\lambda)$, we employ the $l^2(\rho_\lambda)$ error. When evaluating the error of the entire product $g(\lambda) k(\xi)$, we employ the measure $\rho_\xi \rho_\lambda$.


\subsection{Discussion on the large deformation regime} To evaluate the error of $\ub$ on test samples, we solve the learnt model given the test body force $\bb_{test}$ and the Dirichlet boundary condition $\ub(\xb)=\ub_{test}(\xb)$. Similar to its PDE analog, the solution uniqueness results for bond-based peridynamics in Theorem \ref{lemma} only holds for the small deformation regime, i.e., when $ \lambda = \frac{|\xib+\etab|}{|\xib|}\approx 1+\frac{\xib\cdot \etab}{|\xib|^2}$ and $\frac{\xib+\etab}{|\xib+\etab|}\approx\frac{\xib}{|\xib|}$. That means, it is not guaranteed that this uniqueness results hold for materials with large deformation. In fact, the strain energy need not be strictly convex in the large deformation regime, or it will rule out the nonuniqueness essential for the description of buckling \cite{ball1976convexity}. 

However, as will be illustrated further in experiments of Section \ref{sec:convex}, the unique solution in the small deformation regime may serve as a good initialization and thus  help to improve the robustness in the large deformation solution.  
In particular, when employing the learnt MPNO model in down-stream simulation tasks in the large deformation regime, we adopt a two-phase approach to get the solution of $\ub$. We first solve the corresponding small deformation model, i.e.,
\begin{equation}\label{eq:BK_model_linear}
     \int_{B_{\delta}(\bm{0})} g(\lambda)k(\xib) \frac{\xib}{|\xib|} d\xib =- \bb_{test}(\xb,t),
\end{equation}
where the linearized $\lambda = 1+\frac{\xib\cdot \etab}{|\xib|^2}$ is used, and obtain an estimated solution $\tilde{\ub}_{test}$. Then, we employ $\tilde{\ub}_{test}$ as the initial guess for the nonlinear model, and solve for $\ub_{test}$. For both phases, we use the fsolve function from the Python library SciPy (scipy.optimize.fsolve), which employs the Levenberg-Marquardt algorithm. This algorithm iteratively refines the solution by combining the advantages of Newton's method and gradient descent, ensuring robust convergence.

\subsection{Summary of the algorithm}
We summarize the overall algorithm for training and evaluation of the proposed MPNO model in Algorithm \ref{alg:1}. The algorithm comprises two primary components: (1) a training stage, where neural networks are trained to approximate the nonlinear constitutive relation \( g(\lambda) \) and kernel function \( k(\xib) \) while ensuring convexity through architectural constraints, and (2) an evaluation stage, where we evaluate the learning results in three criteria: the error of $\bb$, the error of model, and the error of $\ub$. For the last metric, 
a nonlinear iteration method is employed to solve for $\ub$. 

\begin{algorithm} 
\caption{Learning and evaluation of MPNO.} \label{alg:1}
\begin{algorithmic}[1]
\STATE \textbf{Inputs:} 
    \begin{itemize}
        \item Training, validation, and test datasets $\mcD_{train/valid/test}=\{(\ub^{(i)}_{train/valid/test}(\xb),\bb^{(i)}_{train/valid/test}(\xb))\}_{i=1}^{N_{train/valid/test}}$ on mesh points $\chi = \{\boldsymbol{x}_j\}_{j=1}^{N_x}$.
        \item Horizon size $\delta$ characterizing nonlocal interactions.
    \end{itemize}
\STATE \textbf{Pre-processing data:} 
\STATE \begin{enumerate}
\item Find and store the peridynamic neighbor node set: for each $\xb_j\in\chi$, find all $\xb_k\in\chi$ such that $|\xb_k - \xb_j|< \delta$.
\item If a ground-truth model is given, calculate the empirical measures for $k$ and $g$ following \eqref{eq:L2rho_xi} and \eqref{eq:L2rho_lambda} using the training dataset $\mcD_{train}$.
\end{enumerate}
\STATE \textbf{Training Stage:}
\FOR{$ep = 1: epoch_{max}$}
\FOR{each training batch}
    \STATE \begin{enumerate}
        \item Compute $\xib$, $\etab$, $\lambda$ and loss function \eqref{eq:loss}.
        \item Update parameters $\theta$ via Adam optimizer.
    \end{enumerate}
    \ENDFOR
\FOR{each validation batch}
        \STATE Compute $\xib$, $\etab$, $\lambda$ and the corresponding residual error following \eqref{eqn:Err_b}.
    \ENDFOR
\STATE Store the optimal model parameter $\theta$ if the residual error reduces on the validation dataset $\mcD_{valid}$.
\ENDFOR

\IF{\textbf{Evaluation is desired:}}
\STATE \textbf{Evaluate the residual error:}
\FOR{each test sample $(\boldsymbol{u}_{test}^{(i)}, \boldsymbol{b}_{test}^{(i)})$}
\STATE Compute $\xib$, $\etab$, $\lambda$ and the corresponding residual error following \eqref{eqn:Err_b}.
\ENDFOR
\STATE \textbf{Evaluate the model error:} 
\STATE Compute the kernel errors following \eqref{eqn:Err_kg}.
\STATE \textbf{Evaluate the solution error:}
\FOR{each test sample $(\boldsymbol{u}_{test}^{(i)}, \boldsymbol{b}_{test}^{(i)})$}
    \STATE \begin{enumerate}
    \item Solve the small deformation model \eqref{eq:BK_model_linear} with a zero initial guess of $\ub$, obtain $\tilde{\ub}_{test}$; If small deformation is considered, set $\ub_{test}=\tilde{\ub}_{test}$.
    \item If large deformation case is considered, use $\tilde{\ub}_{test}$ as the initial guess to solve for the large deformation model \eqref{eqn:fullmodel}, and obtain $\ub_{test}$.
    \item Compute the solution error $E_u$ following \eqref{eqn:Err_u}.
    \end{enumerate}
\ENDFOR
\ENDIF
\end{algorithmic}
\end{algorithm}

\section{Verification on synthetic datasets} \label{sec:experiments}
In this section, we verify the proposed framework on a synthetic 1D dataset. In particular, we consider the Blatz-Ko (BK) material model, which is a hyperelastic material model describing material responses like polymer foams \cite{parl1962application,silling2005peridynamic}. In this model, the ground-truth constitutive relation is 
\begin{equation}\label{eq:true_model}
  k(\xi)=k(|\xi|),\,g(\lambda)=\lambda-\lambda^{-3}.
\end{equation}
Here, $k$ represents a radial function that depends on the bond length. 
Specifically, we consider two special examples for $k(\xi)$ in our experiments: 
\begin{itemize}
\item Example I 
\mylabel{ex-i}{(Ex-I)}: $k(\xi) = 2c\cos(\pi|\xi|)$.
\item Example II \mylabel{ex-ii}{(Ex-II)}: $k(\xi) = 2c\exp(-50|\xi|^2)(\delta-|\xi|)$.
\end{itemize}
Here, $c=\mu/\pi/\delta^2$ is a positive constant that is the same in both examples.

\begin{figure}[h!]
    \centering
    \includegraphics[width=0.99\textwidth]{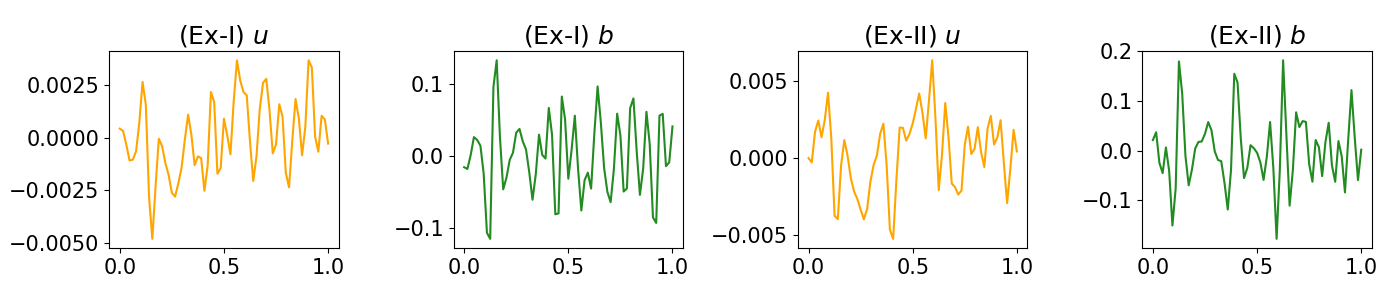}
    \caption{Synthetic dataset from 1D Blatz-Ko model: two exemplar data pairs.
    }
    \label{fig:BK_1d_data} 
\end{figure}

To generate the training, validation, and test datasets, we consider $\omg=[0,1]$, and the following Fourier series representation for the 
displacement $u(x)$:
\begin{equation}\label{eq:analytical_u}
    u^{(i)}(x) = \sum_{j=1}^J (a^{(i)}_j\exp(-j/J)\sin(j\pi x)+b^{(i)}_j\exp(-j/J)\cos(j\pi x)),
\end{equation}
where $J=40$ controls the maximum frequency of the displacement field. The nonlocal interaction radius is chosen as $\delta=0.25$. The corresponding body force $b(x)$ is then numerically generated 
using a refined mesh grid with a spacing of $2^{-8}$. 
To ensure that $\lambda$ is restricted to a physically meaningful range, the coefficients $a^{(i)}_j, b^{(i)}_j$ are randomly sampled from the uniform distribution $\mathcal{U}([-0.001,0.001])$. A total of 400 data samples are generated, from which 300 are randomly selected for training, 50 for validation, and the remaining 50 for testing. For each data sample $u^{(i)}(x)$ and $b^{(i)}(x)$, measurements are provided on a uniform grid with $N_x=257$ points. Fig. \ref{fig:BK_1d_data} illustrate two representative data samples from \ref{ex-i} and \ref{ex-ii}. 

Using this dataset, we aim to demonstrate that the proposed MPNO method can effectively learn the model under all three metrics. 
First, we investigate the convergence behavior of the learning results with respect to the mesh size of the training data. Specifically, we consider the following three learning scenarios:

\begin{enumerate}[label=Case~\arabic*:, ref=\arabic*, align=left, leftmargin=0pt, labelsep=1em, itemindent=6em]
 \item \label{case:1} Assuming $g(\lambda)$ is known, learn $k(\xi)$ only. 
 \item \label{case:2} Assuming $k(\xi)$ is known, learn $g(\lambda)$ only. 
 \item \label{case:3} Learn $g(\lambda)$ and $k(\xi)$ simultaneously. 
\end{enumerate}
Next, we study the importance of obtaining a well-posed model, by comparing the results from MPNO with those obtained using MLPs for both $g$ and $k$. 
To ensure a fair comparison across different experimental settings, we carefully tune hyperparameters, including neural network size, learning rates, and decay rates, to minimize the validation error in $b$. Specifically, the neural network architecture is selected from combinations of 4 or 5 hidden layers, with each layer consisting of 64, 128, or 256 neurons. The initial learning rate is chosen from {0.01, 0.005, 0.001, 0.0005, 0.0001}. The learning rate decay follows an epoch-based schedule, with a decay factor of 0.995 per epoch for the first one-third of the total 50,000 epochs and 0.998 per epoch for the remaining two-thirds.
For the selection of activation functions, we chose from commonly employed options, including Sigmoid, ReLU, Tanh, and Softplus, and pick the one with best performance. In particular, we employ sigmoid for MGN and ReLU for MLP. 
Early stopping is applied based on validation error to prevent overfitting. Optimization is performed using the Adam optimizer. All experiments are conducted on an NVIDIA RTX 3090 GPU with 24 GB of memory.


\subsection{Overall learning results}\label{sec:exp_1d_part1}
\begin{figure}[h!]
    \centering
    \includegraphics[width=0.99\textwidth]{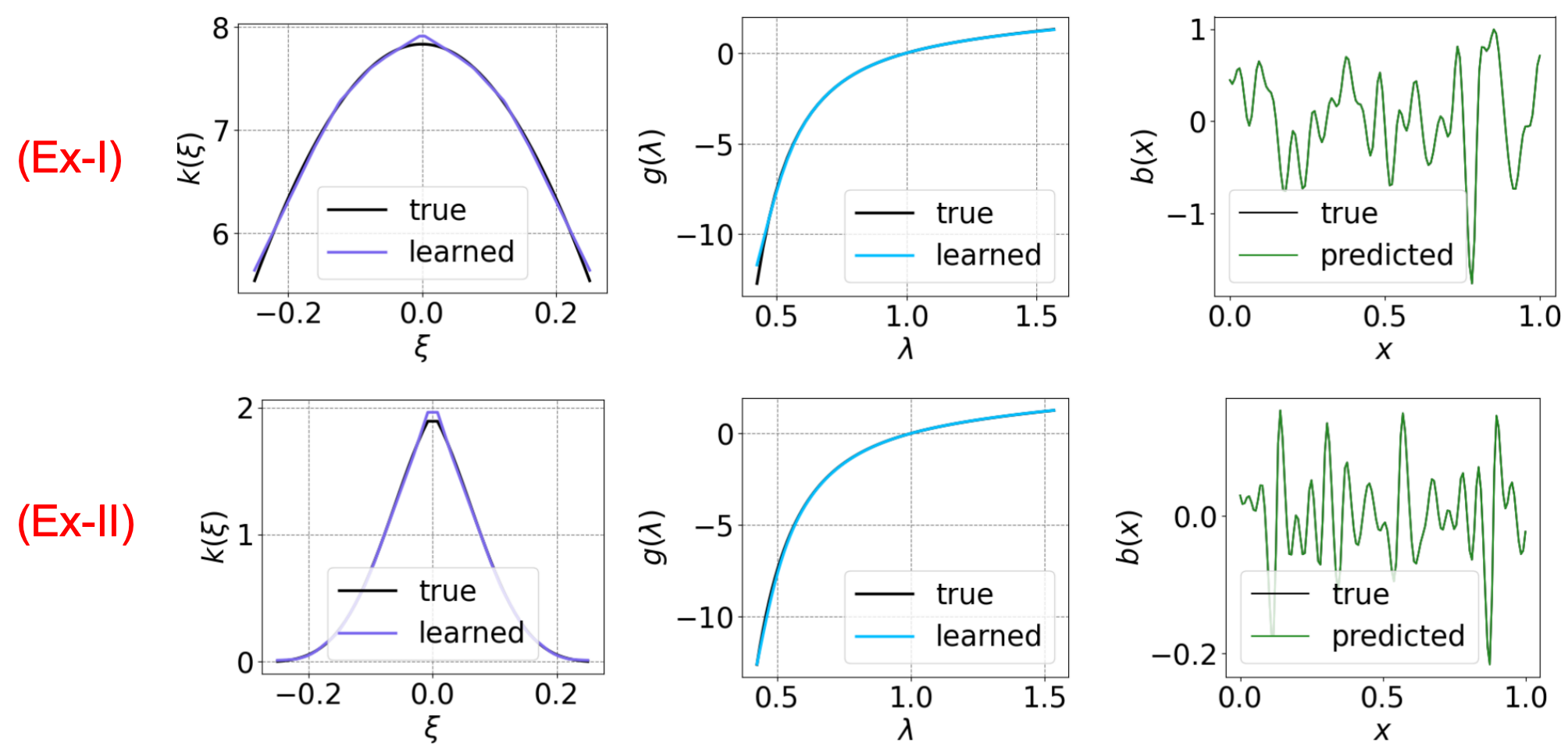}
    \caption{Synthetic dataset from 1D Blatz-Ko model with a fine mesh ($\Delta x=2^{-8}$): results of learning $g(\lambda)k(\xi)$. 
    }
    \label{fig:BK_1d_ex32}
\end{figure}

We begin by evaluating the accuracy of the learned model. In this example, we present the results of simultaneously learning $g(\lambda)$ and $k(\xi)$ using data corresponding to a mesh size of $\Delta x=2^{-8}$. The learned functions $g(\lambda)$ and $k(\xi)$ are plotted in Fig. \ref{fig:BK_1d_ex32} for both two examples. Here, we note that $g$ and $k$ are equivalent with respect to a constant ratio, i.e., $Cg(\lambda)$ and $k(\xi)/C$ would generate the same model as $g(\lambda)$ and $k(\xi)$. {To ensure the identifiability, we normalize the learnt $g^{NN}(\lambda)$ and $k^{NN}(\xi)$ to guarantee that $\int_{B_\delta(0)}k^{NN}(\xi) d\xi=\int_{B_\delta(0)}k(\xi) d\xi$.}
As shown in Fig. \ref{fig:BK_1d_ex32}, the learned functions closely align with the ground-truth. Specifically, the relative $l^2(\rho)$ error of learned $g(\lambda)k(\xi)$ for the two examples are 0.613\% and 0.737\%, respectively.
Additionally, we compare the predicted $b$ with its true counterpart. The average relative $l^2$ error across all 50 test samples is 0.161\% for \ref{ex-i} and 0.016\% for \ref{ex-ii}. A test sample of $b$ is also provided in Fig. \ref{fig:BK_1d_ex32}.

\begin{figure}[h!]
    \centering
    \includegraphics[width=0.99\textwidth]{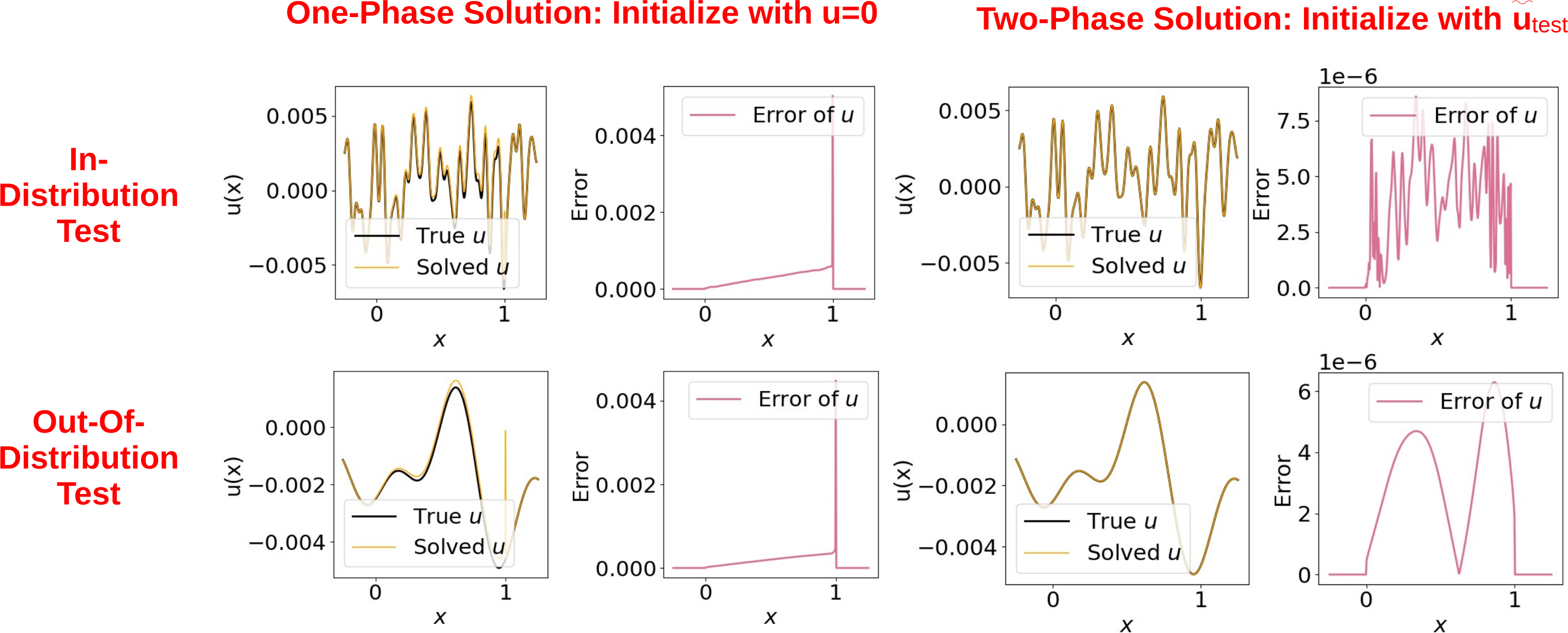}
    \caption{Synthetic dataset from 1D Blatz-Ko model (Ex-I): solving $u$ from the learned model using in-distribution (top) and out-of-distribution (bottom) test datasets. Left: one-phase solution. Right: two-phase solution.}  
    \label{fig:BK_1d_ex22_k:solve_u} 
\end{figure}

Next, we investigate the solution error in downstream simulations using the proposed two-phase strategy. In Fig. \ref{fig:BK_1d_ex22_k:solve_u}, we present representative results from (Ex-I), comparing the solution obtained by our two-phase strategy (referred to as the ``two-phase solution") with that from a direct nonlinear solver without the small deformation initial guess $\tilde{\ub}_{test}$ (referred to as the ``one-phase solution"). As shown in the left-top plot of Fig. \ref{fig:BK_1d_ex22_k:solve_u}, the one-phase method produces reasonable results in most regions but exhibits noticeable discrepancies near the boundaries. In contrast, the two-phase approach yields a more consistent solution that closely matches the true displacement field $u$, see the right-top plot of Fig. \ref{fig:BK_1d_ex22_k:solve_u}.
To quantify the accuracy, we compute the average relative $l^2$ errors over 50 in-distribution test samples for both approaches. For \ref{ex-i}, the one-phase solution results in a higher average error, while the two-phase solution significantly improves the accuracy. Similar trends are observed for \ref{ex-ii}. A summary of all error statistics is provided in Table~\ref{tab:err_BK_1d}.

To further investigate the generalization capability of the learned model, we also consider a set of out-of-distribution (OOD) data pairs $\{(u_{test,OOD}^{(i)}, b_{test,OOD}^{(i)})\}_{i=1}^{N_{test,OOD}}$, generated from lower frequencies $J=5$ in \eqref{eq:analytical_u}, while keeping the true model (Ex-I) unchanged. An exemplar solution of $u$, obtained using both one-phase and two-phase solvers, is presented in the bottom row of Fig. \ref{fig:BK_1d_ex22_k:solve_u}. It is also evident that the initialization strategy using small deformation solution in our proposed two-phase solver substantially improves accuracy, as shown in Table \ref{tab:err_BK_1d}.

\begin{table}[h!]
\caption{Synthetic dataset from 1D Blatz-Ko model: averaged solution errors.}
 \label{tab:err_BK_1d}
\centering
\begin{tabular}{|c|c|c|c|c|}
\hline
 \multirow{2}{*}{errors} & \multicolumn{2}{c|}{In-distribution test} & \multicolumn{2}{c|}{Out-of-distrbution test} \\
\cline{2-5} & one-phase & two-phase  & one-phase  & two-phase \\
\hline
\ref{ex-i}& 1.01\% & 0.168\% & 1.10\% & 0.373\% \\
\hline
\ref{ex-ii} & 1.00\% & 0.0517\% & 0.0184\% & 0.0184\% \\
\hline
\end{tabular}
\end{table}

\subsection{Study on model convergence with respect to the mesh size}
\begin{figure}[h!]
    \centering
    \includegraphics[width=0.24\textwidth]{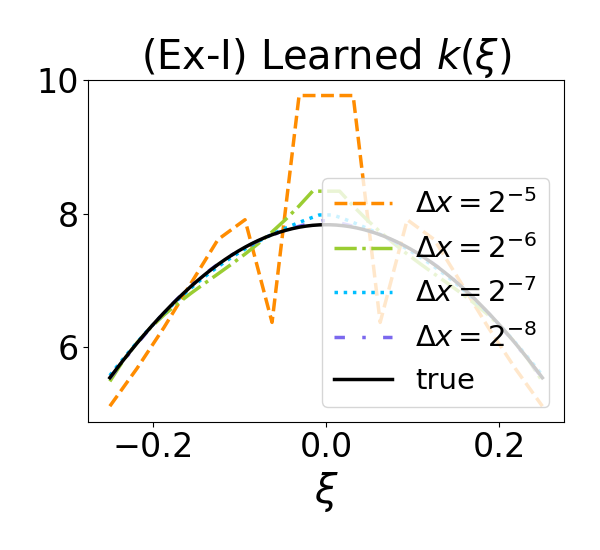}
    \includegraphics[width=0.24\textwidth]{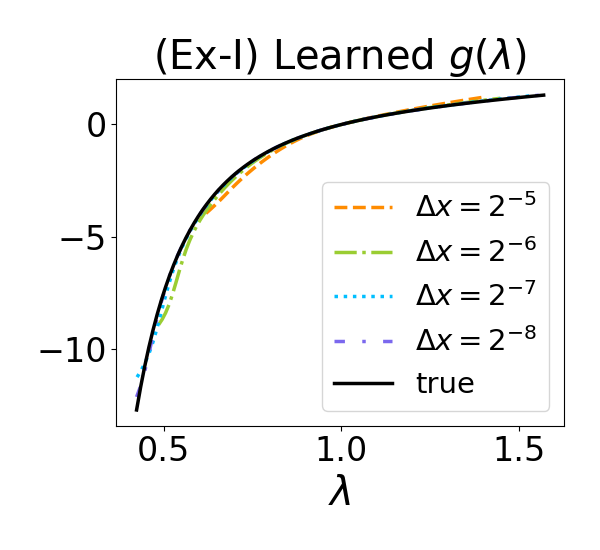}
    \includegraphics[width=0.24\textwidth]{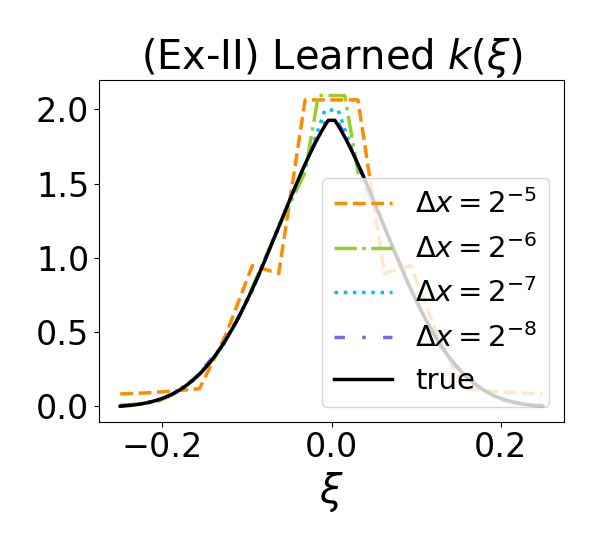}
    \includegraphics[width=0.24\textwidth]{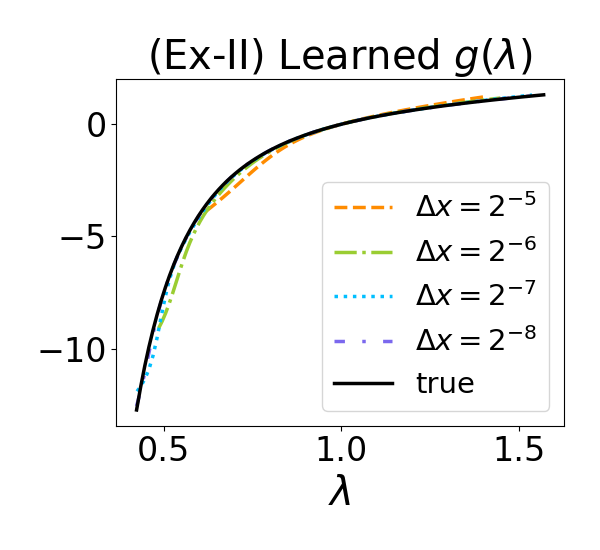}
    \caption{Synthetic dataset from 1D Blatz-Ko model (Ex-I): learned $k(\xi)$ in Case 1 and $g(\lambda)$ in Case 2, as we refine the mesh size $\Delta x$.}
    \label{fig:BK_1d_g_k} 
\end{figure}

\begin{figure}[h!]
    \centering
   \includegraphics[width=0.95\textwidth]{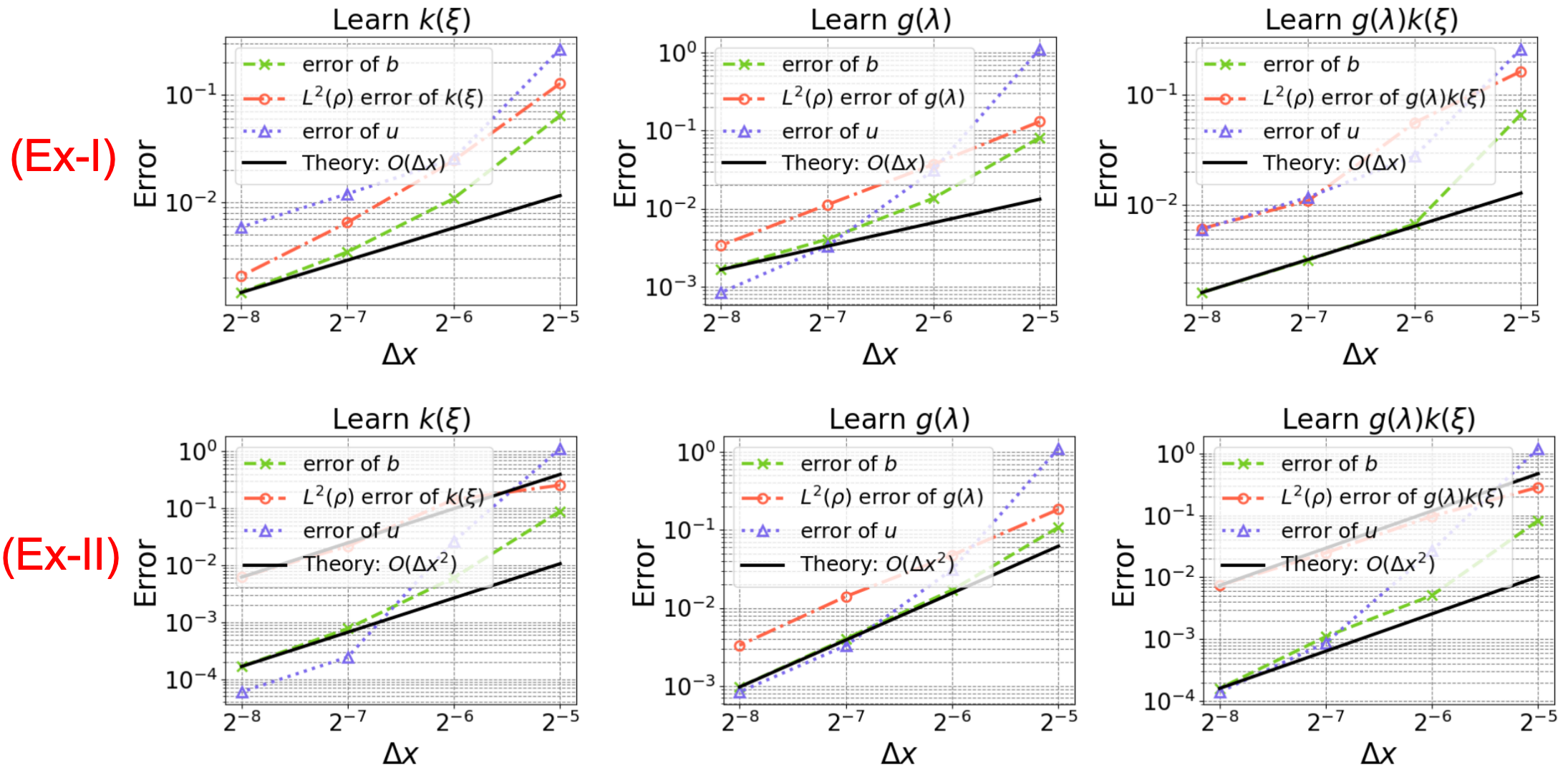}
    \caption{Synthetic dataset from 1D Blatz-Ko model: convergence of the errors of $b$, model and $u$ under three learning cases. 
   }
    \label{fig:BK_1d_case1_order} 
\end{figure}

In this section, we investigate the convergence of model errors.
To examine the impact of different mesh sizes on the learning outcomes, we subsample the training data using mesh sizes $\Delta x=[2^{-5}, 2^{-6}, 2^{-7}, 2^{-8}]$. The model is trained on these four datasets under three different learning scenarios as detailed in Cases~\ref{case:1}–\ref{case:3}.
Fig. \ref{fig:BK_1d_g_k} illustrate the learned functions $k(\xi)$ and $g(\lambda)$ for both \ref{ex-i} and \ref{ex-ii}. As the mesh size becomes finer, the model exhibits reduced errors, indicating improved accuracy. We also quantitatively evaluate the errors of the learned model under all three criteria. These errors are plotted in the top and bottom rows of Fig. \ref{fig:BK_1d_case1_order} for \ref{ex-i} and \ref{ex-ii}, respectively. The results demonstrate a clear first-order convergence for \ref{ex-i} and a second-order convergence for \ref{ex-ii}. In the following, we provide a mathematical analysis of the observed convergence rates.


\textbf{For the Case \ref{case:1} of learning $k(\xi)$ with known $g(\lambda)$:} First, we introduce two assumptions on the solution $u(x)$, model functions $g(\lambda)$ and $k(\xi)$:
\begin{itemize}
\item \mylabel{ass:assump-1}{Assumption 1}: $u \in C^1(\Omega)$, $g\in C^1((0,+\infty)), k\in C^1(B_{\delta}(0)\backslash\{0\})$. 
\item \mylabel{ass:assump-2}{Assumption 2}: $u \in C^2(\Omega)$, $g\in C^2((0,+\infty)), k\in C^2(B_{\delta}(0)\backslash\{0\})$, $ k(\pm \delta)= 0$.
\end{itemize}

Denoting $\xi_l=l\Delta x, ~ l=-L,-L+1,\cdots, L, ~L = \lfloor\delta/\Delta x\rfloor$. 
Under \ref{ass:assump-1}, the integrand function $g(\lambda)k(\xi)\frac{\xi+\eta}{|\xi+\eta|}$ in Eq. \eqref{eq:model} is continuously differentiable about the variable $\xi$, then the Riemann summation approximation yields a numerical error of $O(\Delta x)$. Additionally, the Riemann summation coincides with the trapezoidal rule under the \ref{ass:assump-2}, hence we achieve second-order accuracy. 
That means, the numerical evaluation of $\mcG[u]$ in \eqref{eqn:Riemann_sum} satisfies:
\begin{eqnarray}
\nonumber \mcG_{\Delta x}[u^{(i)}](x_j) &=& \Delta x\sum_{l=-L+1}^L g(\lambda^{(i)}_{j,l})\frac{\xi_l+\eta^{(i)}_{j,l}}{\left|\xi_l+\eta^{(i)}_{j,l}\right|} k(\xi_l)= \mcG[u^{(i)}](x_j)+ \mathcal{O}((\Delta x)^q), 
\label{eq:con_k_riemann}
\end{eqnarray}
where $q=1$ if \ref{ass:assump-1} is satisfied, while $q=2$ if \ref{ass:assump-2} is satisfied.
Note that the training loss can be reformulated as
\begin{eqnarray}\label{eq:loss_k}
\text{loss} 
&=&\frac{1}{N}\frac{1}{N_x}\sum_{i=1}^N\sum_{j=1}^{Nx}\left(\Delta x\sum_{l=-L+1}^L g(\lambda^{(i)}_{j,l})\frac{\xi_l+\eta^{(i)}_{j,l}}{\left|\xi_l+\eta^{(i)}_{j,l}\right|}k(\xi_l)+b^{(i)}(x_j)\right)^2,
\end{eqnarray}
\begin{eqnarray*}
\text{or, equivalently, }\quad \text{loss} = \|A\kb+\bb\|^2_{l^2},
\end{eqnarray*}
where $A\in \R^{(NNx)\times (2L+1)}$ is given as
\begin{eqnarray} \label{eq:A}
A^{(i)}_{j,l}= \frac{\Delta x}{N N_x}g(\lambda^{(i)}_{j,l})\frac{\xi_l+\eta^{(i)}_{j,l}}{|\xi_l+\eta^{(i)}_{j,l}|}, \, A=\left[\begin{array}{cccc}
A^{(1)}_{1,-L}&A^{(1)}_{1,-L+1}&\cdots&A^{(1)}_{1,L}\\
A^{(1)}_{2,-L}&A^{(1)}_{2,-L+1}&\cdots&A^{(1)}_{2,L}\\
\vdots&\vdots&&\vdots\\
A^{(1)}_{N_x,-L}&A^{(1)}_{N_x,-L+1}&\cdots&A^{(1)}_{N_x,L}\\
A^{(2)}_{1,-L}&A^{(2)}_{1,-L+1}&\cdots&A^{(2)}_{1,L}\\
\vdots&\vdots&&\vdots\\
\end{array}\right].
\end{eqnarray}
Correspondingly, $\kb$, $\bb$ are aligned as
\[\kb=[k(\xi_{-L}), \cdots, k(\xi_{L})]^T, \quad \bb=[b^{(1)}(x_1),b^{(1)}(x_2),\cdots, b^{(N)}(x_{Nx})]^T.
\]


Then we have the following truncation error estimate for learning $k(\xi)$.
\begin{lemma}\label{lemma:trun_k}
Let $\kb^{true}$ be the ground-truth discretized kernel and $\kb^{NN}$ be the learned discretized kernel obtained through the training process described in (\ref{eq:loss_k}). Then, the following result holds:
\begin{eqnarray}\label{eq:trun_k}
\|A(\kb^{NN} - \kb^{true})\|_{l^2} = \mathcal{O}((\Delta x)^q) + \mathcal{O}(\|\eb_k\|_{l^2}), 
\end{eqnarray}
where $\eb_k = A\kb^{NN}+\bb$ and $\Delta x$ is the discretization mesh size, $q=1$ if \ref{ass:assump-1} is satisfied; $q=2$ if \ref{ass:assump-2} is satisfied.
\end{lemma}
%
\begin{proof}  
From \eqref{eq:con_k_riemann} one can see that the residual of the ground-truth kernel satisfies
\begin{eqnarray} \label{eq:k_ture}
\rb_k:=A\kb^{true}+\bb= \mathcal{O}((\Delta x)^q).
\end{eqnarray}
Combining $A\kb^{NN}+\bb = \eb_k$, we have the error equation
\[ A(\kb^{NN}-\kb^{true})=-\rb_k+\eb_k.\]
Then the desire conclusion follows.
\end{proof}

\begin{remark}[Error estimate of $k(\xi)$]\label{remark:con_k} Based on the continuity assumption of $u$ and $g$, the matrix $A$ is bounded independently of $\Delta x$. If, in addition, there exists a constant $c>0$, s.t., $\|A^\top A\|_2\ge c$, then $\|(A^\top A)^{-1}A\|_2$ is bounded. Combined with the truncation error estimate \eqref{eq:trun_k}, we obtain the following result for the kernel $k$:
\begin{eqnarray*} \label{eq:con_k}
\|\kb^{NN} - \kb^{true}\|_{l^2} = \mathcal{O}((\Delta x)^q) + \mathcal{O}(\|\eb_k\|_{l^2}),
\end{eqnarray*}
which was what we observed in Fig. \ref{fig:BK_1d_case1_order}.

However, with a fixed training dataset, it was known that the condition number of $A^\top A$ goes to infinity with $\Delta x\rightarrow 0$  \cite{lu2022nonparametric}. Hence, in practice, one may want to make the training dataset more representative, for instance by including higher-frequency components, to mitigate the growth of the condition number of $A^\top A$. 
A more rigorous analysis of the relationship between the representational capacity of the dataset and the convergence of the kernel function is left for future work.
\end{remark}


\textbf{For the Case \ref{case:2} of learning $g(\lambda)$ with known $k(\xi)$:} We can follow the analysis for Case \ref{case:1} to explain its convergence. We divide the range of $\lambda$ into $N_{\lambda}$ intervals: $I_n = [\lambda_n, \lambda_{n+1}]$ for $n=0,1,\dots,N_{\lambda}-1$ and define midpoints $\bar{\lambda}_n = (\lambda_n + \lambda_{n+1})/2$. For each $b^{(i)}(x_j)$, we further approximate the summation in \eqref{eq:con_k_riemann} by summing over values of $\lambda^{(i)}_{j,l} \in I_n$, and approximate $g(\lambda^{(i)}_{j,l})$ by $g(\bar{\lambda}_n)$. This leads to the reformulation:
\begin{equation}\label{eq:con_g_app}
\mcG_{\Delta x}[u^{(i)}](x_j) = 
\sum_{n=0}^{N_{\lambda}-1} B_{j,n}^{(i)}(g(\bar{\lambda}_{n}) + \O(\Delta \lambda)) + \O((\Delta x)^q),
\end{equation}
where
\begin{equation}
B_{j,n}^{(i)} := \Delta x \sum_{l:\lambda^{(i)}_{j,l}\in I_n} k(\xi_l)\frac{\xi_l+\eta^{(i)}_{j,l}}{\left|\xi_l+\eta^{(i)}_{j,l}\right|}.
\end{equation}
Then the training loss can be reformulated as $\|B\gb+\bb\|^2_{l^2}$. Denote
\[ \eb_g := B\gb^{NN}+\bb, \quad \rb_g := B\gb^{true}+\bb.\]
If $B$ is bounded and $\Delta \lambda$ is sufficiently small, then $\|\rb_g\|_{l^2} = \mathcal{O}((\Delta x)^q)$. 
We obtain the following truncation error estimate
\[\|B(\gb^{NN}-\gb^{true})\|_{l^2} = \mathcal{O}((\Delta x)^q) + \mathcal{O}(\|\eb_g\|_{l^2}).\]
Furthermore, similar to the analysis in Remark \ref{remark:con_k}, if $B^\top B$ has a lower bound, then the following result holds
\[\|\gb^{NN}-\gb^{true}\|_{l^2} = \mathcal{O}((\Delta x)^q) + \mathcal{O}(\|\eb_g\|_{l^2}).\]
However, we remark that establishing a lower bound for $B^\top B$ involves a more intricate analysis, as it depends heavily on the structure and representation of the dataset.

Regarding the simultaneous learning of functions $g$ and $k$, a rigorous analysis of the convergence order remains challenging due to the complex coupling between the two functions. We leave a detailed investigation of this issue to future work.

\subsection{Monotone gradient network improves robustness}\label{sec:convex}
\begin{figure}[h!]
\captionsetup{skip=2pt}
    \centering
    \includegraphics[width=0.35\textwidth]{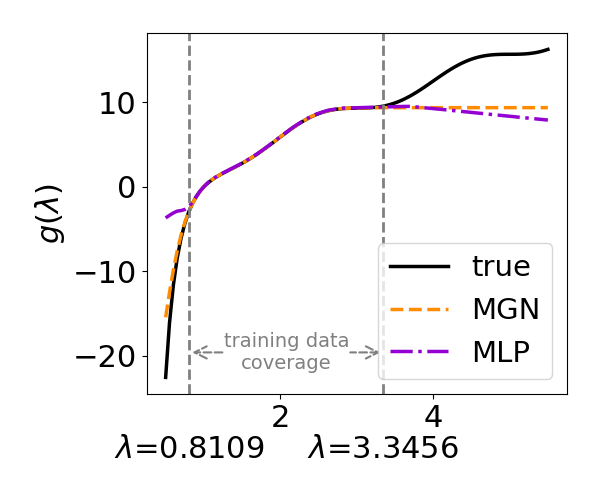}\quad
    \includegraphics[width=0.35\textwidth]{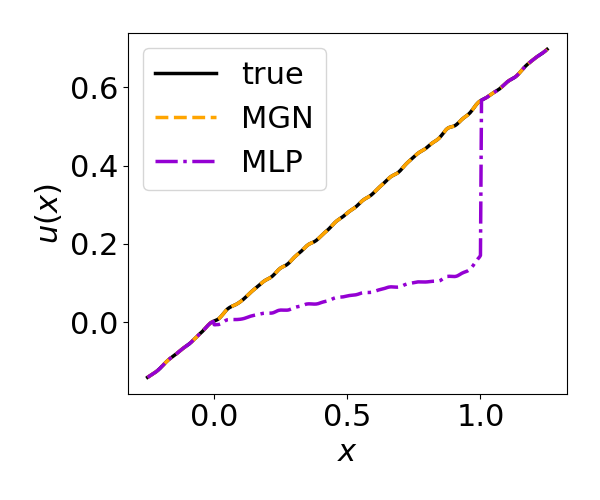}
    \caption{Synthetic dataset from 1D Blatz-Ko model: comparison of learned $g(\lambda)$ using MGN and MLP. Left: learnt models. Right: solutions on an exemplar sample.}
    \label{fig:BK_1d_ex35}
\end{figure}

Theorem \ref{lemma} states that the monotonicity of $g(\lambda)$ ensures uniqueness in the small deformation regime, and our proposed algorithm enforces this feature by employing an MGN while learning $g$. As shown in Fig. \ref{fig:BK_1d_ex22_k:solve_u}, the unique solution provides better robustness both in the small and large deformation regimes. To further verify the advantage of employing an MGN, we also consider a multi-layer perceptron (MLP) architecture for $g$, which does not guarantee convexity, and compare the results with our MPNO using an MGN architecture. 
In particular, we consider a Case 2 setting (learning $g$ only), and generate data with a manufactured ground-truth function $g(\lambda)$ as $
g(\lambda) = \pi(\lambda - \lambda^{-3}) + \sin(\pi\lambda),
$
while the true $k(\xi)=2c\exp(-50|\xi|^2)(\delta-|\xi|)$ is given. To control the range of the bond stretch $\lambda $ in the training data, we add a linear term $C^{(i)} x$  to the analytical solution $u$ in \eqref{eq:analytical_u}, i.e.,
\begin{equation}\label{eq:analytical_u1}
    u^{(i)}(x) = \sum_{j=1}^J (a^{(i)}_j\exp(-j/J)\sin(j\pi x)+b^{(i)}_j\exp(-j/J)\cos(j\pi x))+C^{(i)} x,
\end{equation}
where $J=40$, $a^{(i)}_j,b^{(i)}_j\sim\mathcal{U}([-0.001,0.001])$, and the constant $C^{(i)}\sim\mathcal{U}([0.2,2])$. We use data with a mesh size of $ \Delta x= 2^{-8} $ to train the model.  
Both the MGN and MLP architectures consist of five layers, each with a width of 128. The total number of parameters for MGN and MLP are 1,153 and 66,433, respectively. Due to the shared weights in MGN, it requires a significantly smaller number parameters compared to MLP while still achieving competitive performance.
The learned $g(\lambda)$ obtained by MGN and MLP is illustrated on the left side of Fig. \ref{fig:BK_1d_ex35}. 
The training dataset covers bond stretch values $\lambda$ ranging from 0.8109 to 3.3456. Within this range, both models provide good approximations, closely following the expected function behavior. However, when extrapolating beyond the training domain, we observe a noticeable discrepancy, as expected due to the inherent OOD issue in data-driven learning algorithms. Notably, in the extrapolation domain, the function $g(\lambda)$ learned by MGN preserves its monotonicity, whereas the function learned by the MLP does not necessarily maintain this property. 

To further assess the robustness of the learned models, we evaluate them on all three criteria, and report the averaged errors on 50 test samples. For the solution errors, we considered the two-phase algorithm for both MGN-based and MLP-based models. 
For the MLP-based model we obtain relative errors of 3.53\%, 0.542\%, and 20.9\%  for $b$, $g$, and $u$, respectively. For our MPNO based on MGN, relative errors of 7.14\%, 0.195\%, and 0.065\%  for $b$, $g$, and $u$ are observed. Here, a larger error of $b$ is observed, possibly due to the fact that MLP has a much larger number of trainable parameters and is more expressive. However, MGN-based model generally performs better in the model error and solution error, highlighting its robustness. 
In fact, for many test samples, the MLP-based model fails to converge to the correct solution, whereas the MGN-based model consistently leads to a successfully converged solution. Solutions on an exemplar test sample are demonstrated in the right plot of Fig. \ref{fig:BK_1d_ex35}.


\section{Applications: learning from molecular dynamics simulations} \label{sec:app}



In this section, we employ the MPNO to learn a continuum homogenized model for a two-dimensional molecular dynamics (MD) system. The atoms are generated in a hexagonal lattice with interatomic spacing $\lambda_0=1.4$.
The MD model uses a pair potential with a strongly nonlinear response. This potential results in an interatomic force between any pair of atoms $\alpha$ and $\beta$ is given by $F^{\beta\alpha}(t)=\tanh{\left(\frac{r^{\beta\alpha}(t)-r^{\beta\alpha}(0)}{d}\right)}$, $r^{\beta\alpha}(0)\le3\lambda_0$, where $d=0.05$ and $r^{\beta\alpha}$ is the current distance between the atoms.
Atoms separated by a distance greater than $3\lambda_0$ do not interact, hence we take the horizon size $\delta=4.2$.
The MD simulations determine the equilibrium atomic displacements under predefined external force fields and boundary conditions. A coarse-graining method is applied to the raw MD data to derive the data format used in our experiments. 
Details on the MD data generation can be found in \cite{silling2023peridynamic,you2022data}.

\begin{figure}[h!]
    \centering
    \includegraphics[width=0.99\textwidth]{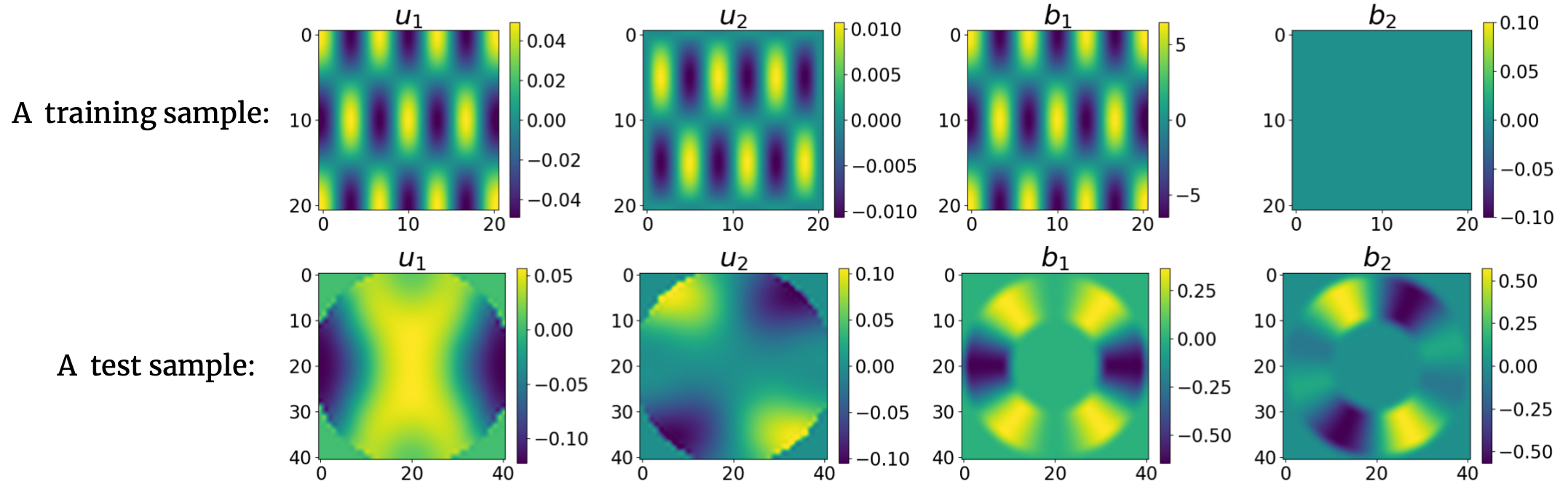}
    \caption{Molecular dynamics simulation dataset: exemplar training and test samples.}
    \label{fig:MD_data} 
\end{figure}

For training and validation, the MD simulation domain consists of a 100$\times$100 square region.  The external loading is applied to the atoms in the MD grid according to an oscillatory function of position, with varying wavelengths for each sample and increasing linearly over time. These forces on each atom $\alpha$ are given by
\begin{eqnarray*}
&&\bm{B^\alpha}=\left(C_{k_1, k_2}^1 \cos \left(k_1 x^\alpha_1\right) \cos \left(k_2 x^\alpha_2\right), 0\right), \text { or } \bm{B^\alpha}=\left(0, C_{k_1, k_2}^2 \cos \left(k_1 x^\alpha_1\right) \cos \left(k_2 x^\alpha_2\right)\right),
\end{eqnarray*}
where $C_{k_1, k_2}^1$ and $C_{k_1, k_2}^2$ are amplitude coefficients, and $k_1$ and $k_2$  are wave numbers corresponding to the spatial frequencies in the $x_1$ and $x_2$ directions, respectively. 
The wave numbers are given by all combinations of $k_1,k_2=\frac{\pi J}{L}, \; J\in\{0,1,2,3,4,5\}$, where $L=100$ is the side length of the square sample. The case $k_1=k_2=0$ is excluded. The amplitude coefficients are chosen so that by the end of each calculation, the system is well into the nonlinear region of the interatomic force dependence. These amplitude coefficients consist of two sets of cases, each of which is evaluated for all $k_1,k_2$ as given above:
\begin{eqnarray*}
 C_{k_1, k_2}^1=81(k_1+k_2), \,C_{k_1, k_2}^2=0,  \text{ or } C_{k_1, k_2}^1=0, \,C_{k_1, k_2}^2=81(k_1+k_2).
\end{eqnarray*}
With all of these combinations of $\{k_1,k_2,C_{k_1, k_2}^1,C_{k_1, k_2}^2\}$,
the dataset comprises 70 distinct loading cases. 20 time records are captured for each case, resulting in a total of 1,400 MD samples. The loading is increased over time according to a slowly increasing ramp function. This loading rate is slow enough that the entire grid is close to equilibrium at all times. 
The loading stops increasing when the maximum strain, defined as $r^{\alpha\beta}(t)/r^{\alpha\beta}(0)-1$, reaches 0.25 for some $\alpha$ and $\beta$ anywhere in the grid. We then only keep the samples before stopping, and obtained 274 samples in total. 
Through a coarse-graining procedure, the MD data is transformed into displacement and force fields discretized on a uniform 21$\times$21 with nodal spacing $h=5$.
A representative data sample, illustrating both displacement and body force, is depicted in the first row of Fig. \ref{fig:MD_data}.

To demonstrate the generalization capability of the constitutive model across different loading and domain settings, we generate a test dataset on a circular domain with radius 100. The loading is applied specifically to an annular region between radii 50 and 100. There are four loading cases, ${\bm{B}}_{test}^{(m)}$, $m=1,2,3,4$, with the force vector on each atom $\alpha$ given by
\begin{align*}
&\bm{B}_{test}^{(1)^\alpha}=\big(2.28(\cos4\theta^\alpha)(\cos\theta^\alpha)r(t),\; 2.28(\cos4\theta^\alpha)(\sin\theta^\alpha)r(t)\big),  \\
&\bm{B}_{test}^{(2)^\alpha}=\big(1.76\,({\mathrm{sign}}(\cos4\theta^\alpha))(\cos\theta^\alpha)r(t),\;1.76\,({\mathrm{sign}}(\cos4\theta^\alpha))(\sin\theta^\alpha)r(t)\big), \\
&\bm{B}_{test}^{(3)^\alpha}=\big(0,\; 0.855\,{\mathrm{sign}}(\sin\theta^\alpha)r(t)\big),  \\
&\bm{B}_{test}^{(4)^\alpha}=\big(1.33\,({\mathrm{sign}}(\sin3\theta^\alpha))(\sin\theta^\alpha)r(t),\;  1.33\,({\mathrm{sign}}(\sin3\theta^\alpha))(\cos\theta^\alpha)r(t)\big),  
\end{align*}
where $\theta^\alpha$ denotes the polar angle of $(x^\alpha_1,x^\alpha_2)$, and $r(t) = t/2000$ is a slowly increasing ramp function.
Functions ${\bm{B}}_{test}^{(m)} (m=2,3,4)$ are discontinuous in the spatial coordinates, creating high-frequency content that provides a challenging test of the method.
Each of the four cases is sampled at 20 time points with a time step of 0.04, resulting in a comprehensive test set of 80 instances. Because of the larger size of the region compared with the region used for training, these test cases employ a 41$\times$41 coarse-grained grid rather than 21$\times$21 (as illustrated in the second row of Fig. \ref{fig:MD_data}).
The spacing between the coarse-grained nodes is the same as in the training cases.

\begin{figure}[h!]
    \centering
    \includegraphics[width=0.9\textwidth]{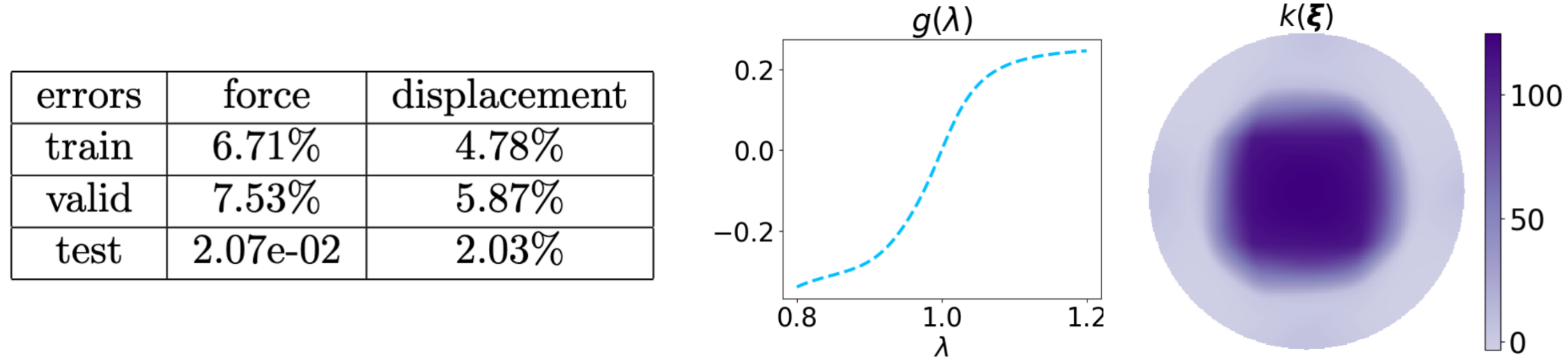}
    \caption{Molecular dynamics simulation dataset: results from MPNO. Left: training, validation and test errors. Right: learned $g(\lambda)$ and $k(\xib)$.}
    \label{fig:MD_g_k_error} 
\end{figure}

\begin{figure}[h!]
    \centering
    \includegraphics[width=0.99\textwidth]{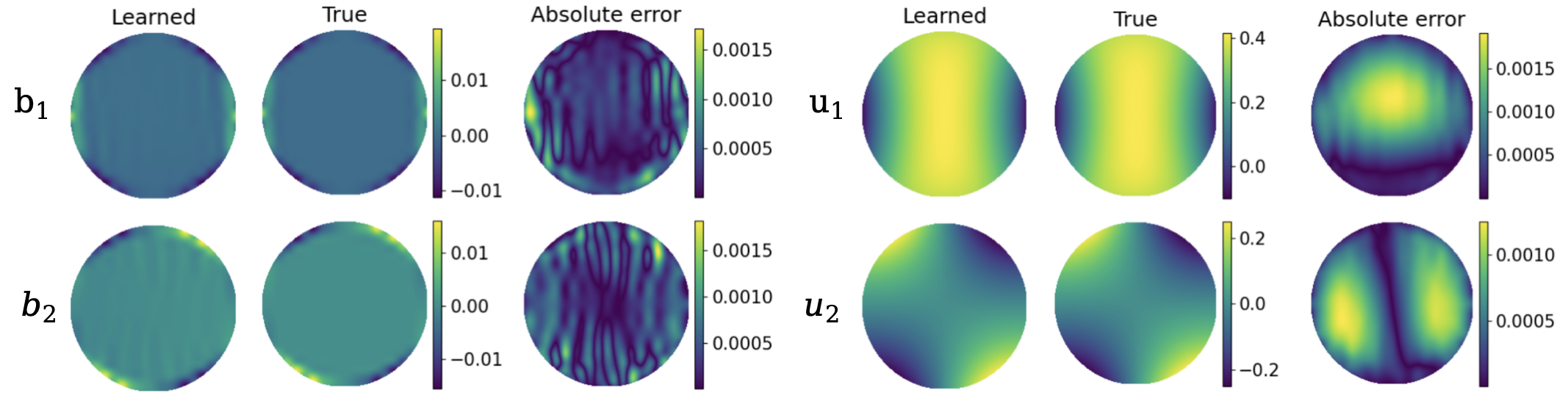}
    \caption{Molecular dynamics simulation dataset: exemplar learning results from MPNO. Left: comparison of the predicted body force $\bb$ versus the corresponding ground truth. Right: comparison of the predicted solution $\ub$ versus the corresponding true displacement.}
    \label{fig:MD_test} 
\end{figure}

During the learning process, we randomly selected 200 samples for training and 50 samples for validation from a total of 274 samples defined on square domains. All 80 samples defined on the circular domain were used for testing.
The MGN used for learning $g(\lambda)$ consists of 16 hidden layers, each with a width of 16, while the MLP for $k(\xib)$ has 4 hidden layers, each with a width of 256. The activation function used in the MGN is $\alpha_l\odot \text{Sigmoid}(x)+ x$ for each layer.
Table in Fig. \ref{fig:MD_g_k_error} reports errors of $\bb$ and $\ub$ for training, validation and test datasets. We report the relative $L^2$ errors, expect for the force error in the test data, where we use the maximum absolute error since $\bb_{test}=\bm{0}$ in the computational domain.
The learned stretch function $g(\lambda)$ and $k(\xib)$ are demonstrated in Fig. \ref{fig:MD_g_k_error}. The range of bond stretch $\lambda$ in training data is approximately $0.89\sim 1.11$. Within this range, the stretch function $g(\lambda)$ exhibits nonlinearity, and the influence function $k(\xib)$ is radially symmetrical, indicating the isotropic property of the material. In Fig. \ref{fig:MD_test}, we plot the predicted body force $\bb$ compared with the true $\bb$ and the displacement $\ub$ solved from the learned model compared with the true $\ub$ for the test data.

\section{Conclusion and future directions}
\label{sec:conclusions}

In this work, we have developed a novel constitutive law learning framework, which we call the Monotone Peridynamic Neural Operator (MPNO). By characterizing a bond-based peridynamic model in the form of a neural operator, MPNO learns a nonlinear material model from spatial measurements of displacement and loading fields. Compared to other data-driven constitutive model learning approaches, the MPNO features a conditional guarantee of model well-posedness. Specifically, we have for the first time provided a sufficient condition of solution uniqueness for nonlinear bond-based peridynamics in the small deformation regime. This condition is encoded in the neural architecture via a monotone gradient network to guarantee the convexity of the resulting strain energy function. In the large deformation regime, we propose a two-phase solution algorithm to enhance the robustness of the solution procedure. To verify our proposed learning workflow and the performance of MPNO in down-stream simulation tasks, we apply it to a synthetic dataset. For this dataset, we investigate the convergences of the residual, the model, and the solution as the measurement grid is refined. At least first order convergence is obtained for all three criteria. The applicability of MPNO is further validated by learning a material model from molecular dynamics (MD) simulations, and then showing satisfactory agreement of solutions with MD results in a much different geometry, under different loading conditions.

Despite these encouraging results, potential extensions merit further study. One such extension is the generalization of the model to guarantee solution existence, analogously to the role of strain energy polyconvexity in classical PDE-based models \cite{ball1976convexity}. If this generalization can be accomplished, then the model could potentially be applied to more complex mechanical phenomena such as buckling and crack propagation \cite{you2023towards}. Another interesting direction would be the extension of the present bond-based peridynamic model to state-based models \cite{silling2007peridynamic}, including a theoretical investigation of well-posedness with this extension. 
Finally, we believe that MPNO can provide a base framework to develop a foundational material model \cite{yu2024nonlocal}, where all materials share the same bond stretch dependence $g$ but have different, and possibly spatially dependent, mechanical properties/microstructures as characterized by different kernels $k$. Such a base framework would, by straightforward extension of our present results, guarantee well-posedness and could capture material-specific heterogeneity. 

\section*{Acknowledgments}

This article has been authored by an employee of National Technology and Engineering Solutions of Sandia, LLC 
under Contract No. DE-NA0003525 with the U.S. Department of Energy (DOE). 
The employee owns all right, title and interest in and to the article and is solely responsible for its contents. 
The United States Government retains and the publisher, by accepting the article for publication, 
acknowledges that the United States Government retains a non-exclusive, paid-up, irrevocable, 
world-wide license to publish or reproduce the published form of this article or allow others to do so, 
for United States Government purposes. 
The DOE will provide public access to these results of federally sponsored research in 
accordance with the DOE Public Access Plan https://www.energy.gov/downloads/doe-public-access-plan.

\end{document}